# A Supervised Adverse Drug Reaction Signalling Framework Imitating Bradford Hill's Causality Considerations

Jenna Marie Reps[a,*], Jonathan M. Garibaldi[a], Uwe Aickelin[a], Jack E. Gibson[b], Richard B. Hubbard[b]

[a]*School of Computer Science, University of Nottingham, NG8 1BB, UK*
[b]*Division of Epidemiology and Public Health, University of Nottingham, UK*

---

## Abstract

Big longitudinal observational medical data potentially hold a wealth of information and have been recognised as potential sources for gaining new drug safety knowledge. Unfortunately there are many complexities and underlying issues when analysing longitudinal observational data. Due to these complexities, existing methods for large-scale detection of negative side effects using observational data all tend to have issues distinguishing between association and causality. New methods that can better discriminate causal and non-causal relationships need to be developed to fully utilise the data.

In this paper we propose using a set of causality considerations developed by the epidemiologist Bradford Hill as a basis for engineering features that enable the application of supervised learning for the problem of detecting negative side effects. The Bradford Hill considerations look at various perspectives of a drug and outcome relationship to determine whether it shows causal traits. We taught a classifier to find patterns within these perspectives and it learned to discriminate between association and causality. The novelty of this research is the combination of supervised learning and Bradford Hill's causality considerations to automate the Bradford Hill's causality assessment.

We evaluated the framework on a drug safety gold standard know as the observational medical outcomes partnership's nonspecified association reference set. The methodology obtained excellent discriminate ability with area under the curves ranging between 0.792-0.940 (existing method optimal: 0.73) and a mean average precision of 0.640 (existing method optimal: 0.141). The proposed features can be calculated efficiently and be readily updated, making the framework suitable for big observational data.

---

## 1. Introduction

Side effects of prescription drugs, also known as adverse drug reactions (ADRs), occur unpredictably and present a major healthcare issue. It is possible that a generally healthy individual may take a prescription drug for a minor problem and end up with a potentially life threatening ADR. As a consequence, it is essential to monitor all marketed drugs and develop methods that are capable of identifying ADRs at the earliest possible point in time. The potential benefits of utilising longitudinal observational data for detecting (also known as signalling) ADRs have been highlighted [1]. However, unsupervised methods developed to signal ADRs using longitudinal observational data have been found to obtain high false positive rates consistently across data sources [2, 3]. This is due to the complexities of observational data, such as missing data and confounding, making it difficult for the methods to distinguish between association and causality. Reference sets detailing known ADRs and non ADRs have been created to aid the development of ADR signalling methods for longitudinal data by enabling a fair evaluation of the methods' ADR signalling performances [4]. However, the creation of reference sets now presents the opportunity of generating labelled data and developing a supervised framework that can be applied to longitudinal observational data to signal ADRs. The success of a supervised framework relies on identifying suitable features for discriminating between causal and noncausal relations. The Bradford Hill causality considerations are a collection of nine factors that are often considered by experts to evaluate whether a drug and health outcome pair may correspond to an ADR [5, 6, 7]. Therefore, the Bradford Hill causality considerations seem an ideal basis for engineering suitable causal discriminative features to be used as input to train an ADR signalling classifier. The aim of this paper is to investigate whether such a classifier can be trained to successfully automate the process of using the Bradford Hill causality considerations to identify causality.

Our proposed supervised Bradford Hill's methodology is evaluated by considering the problem of signalling ADRs that occur shortly after being prescribed a medication. The data used in this study is a large UK electronic healthcare database that contains medical records for millions of patients in the UK. The database is over 300GB in size, therefore it is important to consider the efficiency of the feature engineering. The Bradford Hill's causality considerations were developed by an epidemiologist in the 60s with experience in identifying causal relationships between drugs and health outcomes. They have been successfully implemented, by the process of manual review, as a means to determine causality in many epidemiological studies [8]. The considerations state that nine factors should be considered when assessing causality between a drug and health outcome. The factors are: i) association strength, ii) temporality, iii) consistency, iv) specificity, v) biological gradient, vi) experimentation, vii) analogy, viii) coherence and ix) plausibility. As longitudinal observational databases contain data that can give insight into many of these considerations, we should take advantage of the data available to create a supervised signal detection framework that can imitate the causality review process.

The problem of identifying ADRs, has often relied on the use of spontaneous reporting system (SRS) data. SRS data are composed of reported cases where somebody has suspected that a drug caused an ADR [9]. Common methods for detecting ADRs using SRS data are the disproportionality methods [10] that calculate a measure of association strength between the drug and health outcome based on inferring approximate background rates using all the reports. However, it is not possible to calculate the actual background incidence rates corresponding to the drug or health outcome using SRS data. Issues with under-reporting [11] can limit the ability to detect ADRs using SRS data and consequently, there has been an interest in using longitudinal observational data to aid ADR detection. Recent advances in using SRS data for signalling ADRs have focused on utilising all the SRS data and have considered nonassociation strength features [12, 13]. It was shown that considering a variety of features lead to an improvement in ADR detection compared to standard methods [12]. However, this idea is currently unexplored for ADR detection using longitudinal observational databases, although there has been preliminary work suggesting Bradford Hill based features may add a new perspective for analysing electronic healthcare records [14].

Longitudinal observational data has been a recent focus of attention for extracting new drug safety knowledge due to it being a cheaper and often safer alternative to experimentation such as randomised controlled trials. Existing method for signalling ADRs using longitudinal observational databases include adapted disproportionality methods [15, 16], association rule mining techniques [17, 18], or adaptions of epidemiological studies [19]. All the large scale signalling methods are unsupervised, focus mostly on the measure of association strength and tend to have a high false positive rate in real life data [2, 3], although some supervised techniques have been developed for specific cases. In [20], an ensemble technique combining simple epidemiology study designs to identify paediatric ADRs was shown to perform well. This suggested that incorporating supervised learning for ADR detection might lead to the improvement of signalling ADRs. For supervised learning to be fully utilised in this field, it is important to identify suitable features for the model. This motivates the idea of using a standard set of causal considerations widely implemented by experts in the field of epidemiology as a basis to engineer features. Numerous observational databases, including electronic healthcare records, tend to have hierarchies in the data recording [21, 22]. It may be important to consider the hierarchies when searching for causal relationships because the relationship may be non-obvious when considering a high level item due to it occurring less frequently, but obvious when an abstract perspective is taken. If not taken into consideration, the hierarchal nature of the databases may weaken a signal. Therefore, we also propose features based on medical event coding hierarchies.

Outside of the field of drug safety, existing methods developed with the aim of identifying causal relationships within longitudinal observational data are often based on Bayesian networks [23]. Due to the complexity of creating a complete Bayesian network, many of the proposed methods are considered inappropriate for ‘big’ data [24]. However, constraint-based causal detection has been suggested as a means to handle ‘big’ data by applying metaheaurisics that reduce the problem space [25]. Unfortunately these methods cannot overcome the common issues found within medical longitudinal data such as selection bias and do not consider hierarchal structures, and are therefore not currently suitable for signalling ADRs.

The continuation of this paper is as follows. Section 2 details the database used within this research and the proposed supervised Bradford Hill framework. In section 3 we present the results of the supervised Bradford Hill framework’s performance for signalling ADRs using a real database containing millions of UK patient records. The implications of the results are discussed in section 4. The paper concludes with section 5.

## 2. Materials and Methods

### *2.1. THIN Database*

The data used in this paper were extracted from The Health Improvement Network (THIN) database, an electronic healthcare database containing UK primary care records for over 3.7 million active patients [26] (www.thin-uk.com). As the database contains time stamped records of medical events (e.g., myocardial infarction or vomiting) and drug prescriptions, each patient's medical state can be observed over time and temporal relationships between drugs and medical events can be identified. The THIN data used in this research contained over 200 million medical records and over 350 million prescription records

The THIN database consists of heterogeneous data with multiple hierarchal structures. The database contains three key tables; the patient table, the medical table and the therapy table. For privacy reasons the patients' identities are not stored in the database, instead, each patient is assigned a unique reference known as the patientID that is used to determine which patient each record in the database corresponds to. The patient table contains information about each patient such as their date of birth, gender and date of registration or date of death (if they have died). The medical and therapy tables contain time stamped records of any medical or therapy event experienced by the patients, respectively. The database is normalised such that medical event descriptions and drug details are stored into separate tables and linked with unique references. The unique reference of a medical event is known as the Read code [22] and the unique reference of a drug is known as a drugcode.

The Read codes have a hierarchical coding system encompassing five levels of specificity, with level one Read codes representing very general events and level five Read codes representing very specific events. The level of a Read code is determined by its length. An example of a level one Read code is '1' and an example of a level 5 Read code is '11a1b'. The level 1 Read code 'G' is the parent of any Read code starting with 'G'. For example, the level 1 Read code 'G' representing the medical event 'Circulatory system disease', it is the parent of the Read codes:

Level 2 : 'G5' - 'Other forms of heart disease'

Level 3 : 'G57' - 'Cardiac dysrhythmias'

Level 4 : 'G57y' - 'Other cardiac dysrhythmias'

Level 5 : 'G57y1' - 'Severe sinus bradycardia'

We define an equivalence relationship between Read codes as, $Readcode_m$ ⍰ $Readcode_n$ if the level $k$ parent of

2

$Readcode_m$ is the same as the level $k$ parent of $Readcode_n$. For example, $G$51 ⍰ $G$5724. Prescription drugs are recorded via a drugcode and have an associated British National Formula (BNF) code [27]. The BNF code also has a hierarchal structure and can be used to identify similar drugs. In the THIN database there are more than 66,000 drugcodes and 100,000 Read codes.

#### *2.1.1. THIN Processing*

Care needs to be taken with newly registered patients as patients can move to new general practices at any point in their life and may have existing medical conditions that get recorded when they join the new practice. Newly registered patients have the potential to bias results as doctors may record medical events that are pre-existing, but these events will have an incorrect timestamp. It has been shown that the probability of pre-existing medical events being recorded into the THIN database for newly registered patients is significantly reduced after a year of the patient being registered [28]. Consequently, to prevent newly registered patients biasing results, the first year after registration is ignored in this study. We also ignore drug prescription records that occurred within the final month of the latest THIN data collection date as including these might cause under-reporting.

### *2.2. Supervised ADR signalling Framework*

The supervised ADR framework (SADR) involves three steps. Figure 1 presents the flowchart of the framework, showing how the framework is used to signal new ADRs once a classifier is trained. The first step is to extract the dataset based on existing reference sets developed for evaluating signal detection methods for longitudinal observational data [29]. This step involves identifying the drug-health outcome pairs and their label (e.g., is the drug a known ADR or non-ADR of the health outcome?). The second step of the SADR framework is to implement feature engineering to combine existing drug safety measure features and create novel features that cover the Bradford Hill strength, temporality, specificity, experimentation and biological gradient considerations. The other considerations were not included due to the information not generally being available in a single longitudinal observational database. Features specific to the hierarchal structures of medical event codings are also proposed. The third step is training a binary classifier using the reference set labels and Bradford Hill consideration based features to produce a supervised ADR signal detection model that can discriminate between causality and association. To evaluate the suitability of the various features we will train the classifier on a subset of the labelled data and validate the classifier on the remaining subset of the labelled data to investigate the agreement between the prediction and truth and ensure the classifier is generalizable (i.e., should perform as well on new data). Due to limited definitive knowledge of known ADRs and non-ADRs, this is the best approach to determine the performance of the classifier for signalling ADRs. In this paper we use the random forest binary classifier as this was shown to perform well for the THIN data in previous work [20] and this was also supported by preliminary results.

The SADR framework is only applied to drug and Read code pairs where the Read code occurred for 3 or more patients within the month of the drug prescription. This restriction is due to this paper focusing on ADRs that occur shortly after the drug is ingested, so a month period is a trade off between ensuring the ADR is recorded while reducing the amount of noise. If the Read code occurs for less than 3 patients within the month after the drug then it would be very difficult to statistically show the corresponding medical event is an ADR and the three or more limit is a common threshold applied in pharmacovigilance [30].

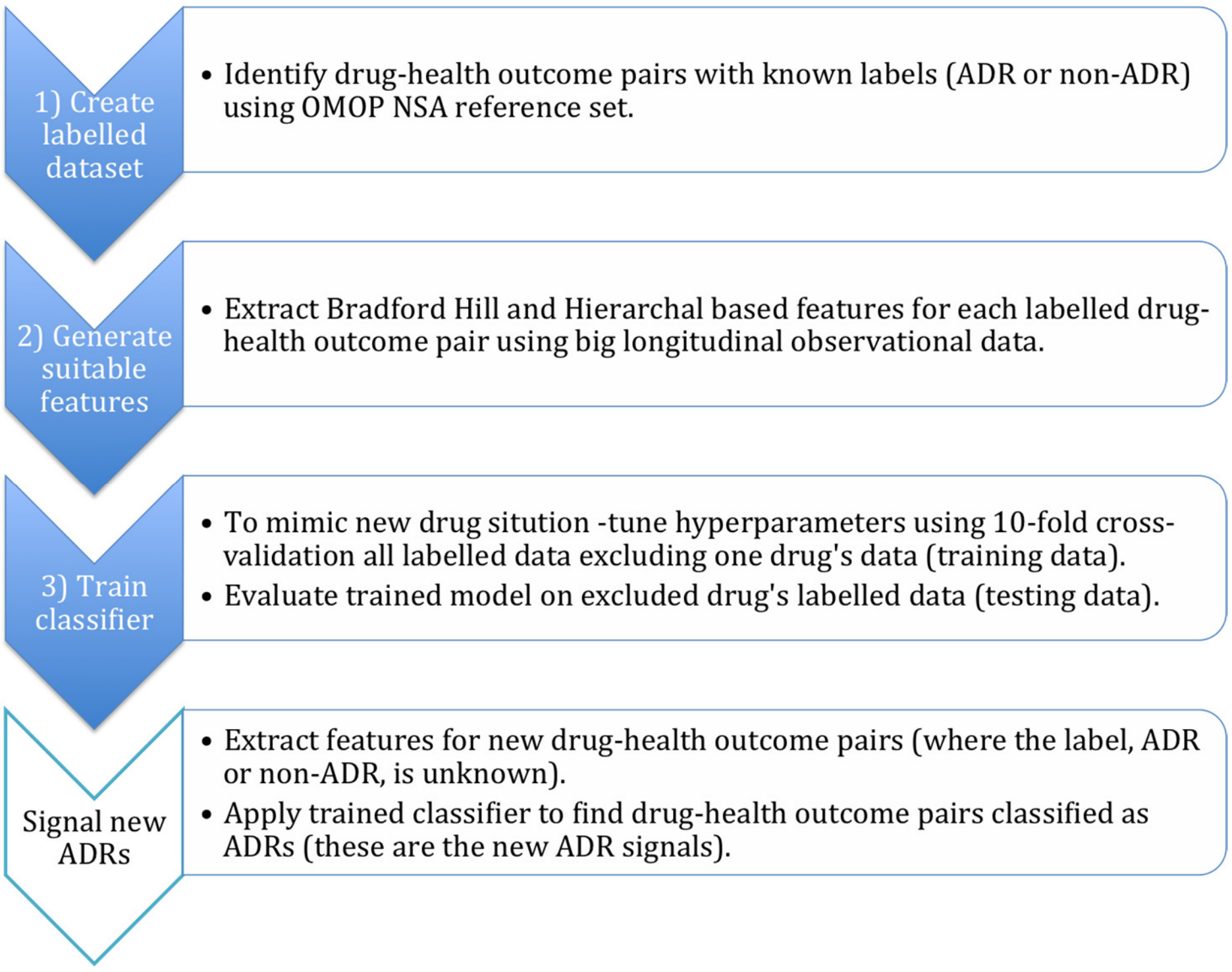

Figure 1: The overall flowchart of the SADR framework

*2.2.1. Step 1: Create Dataset*

The observational medical outcomes partnership (OMOP) have provided a non-specified association (NSA) reference set containing drug-health outcome of interest (HOI) pairs known to correspond to ADRs or non-ADRs [29]. This reference set was generated specifically for evaluating signal generating methods to enable a fair comparison. This set states whether the HOI occurs shortly after drug exposure, or after long term exposure.

Matching the OMOP NSA reference set with the THIN data caused some issues. THIN record their prescriptions using a unique drug coding system (a transformation of the multilex code) and there is currently no way to map the THIN prescription coding to the OMOP reference set coding without requiring extensive manual work from the THIN staff. To overcome this issue we managed to identify suitable BNF codes for the OMOP drugs and used the BNF as a way to match the OMOP drug and the THIN drugcodes. There is a mapping between the THIN Read codes and the OMOP health outcome codes, however this is not one-to-one as the Read clinical coding suffers from redundancy. Consequently, it was common to have multiple drug-Read code pairs for each drug-HOI pair.

There were issues finding a suitable BNF code for Amphotericin. We could not find any prescriptions of drugs with the BNF code 05020300 (Amphotericin) in the THIN data as Amphotericin was recorded with the BNF codes 05020000, 13100200 or 12030200 but these also corresponded to non-Amphotericin drugs. Therefore we excluded

Amphotericin from the analysis. Table 1 shows the drugs' BNF mappings used and the number of HOI or Read codes paired with each drug in the dataset. It can be seen there were multiple Read codes for each HOI. As the performance of a classifier is likely to improve with a larger training set, we decided to keep the drug-Read code pairs rather than aggregating them into drug-HOI pairs, although this does means some of the signals may be weaker.

Table 1: Mapping between OMOP drugs and BNF codes used to map OMOP drugs to THIN drugs.

| | | Count | |
|---|---|---|---|
| Drug | BNF | HOI | Read |
| OMOP ACE Inhibitor | 02.05.05.01 | 527 | 566 |
| OMOP Antibiotics | 05.01.08.00 | 292 | 301 |
| OMOP Antibiotics | 05.01.03.00 | 232 | 243 |
| OMOP Antibiotics | 05.01.05.00 | 299 | 315 |
| OMOP Antiepileptics | 04.08.01.00 | 417 | 440 |
| OMOP Benzodiazepines | 04.01.02.00 | 345 | 371 |
| OMOP Benzodiazepines | 04.08.02.00 | 6 | 6 |
| OMOP Beta blockers | 02.04.00.00 | 499 | 528 |
| OMOP Bisphosphonates | 06.06.02.00 | 282 | 301 |
| OMOP Tricyclic antidepressants | 04.03.01.00 | 443 | 478 |
| OMOP Typical antipsychotics | 04.02.01.00 | 307 | 332 |
| OMOP Typical antipsychotics | 04.06.00.00 | 7 | 7 |
| OMOP Typical antipsychotics | 23.00.00.00 | 1 | 1 |
| OMOP Warfarin | 02.08.02.00 | 327 | 345 |

At this point we had a list of 4249 drug-Read code pairs, *$drug_i$-$Readcode_j$*, and their ground truth labels (e.g., whether the pair is a known ADR or non-ADR). The next step was to generate suitable features for each pair.

### *2.2.2. Step 2: Features Engineering*

Features based on the five factors of the Bradford Hill considerations and specific to the hierarchal medical event coding structures were calculated per drug-Read code pair. A summary of these features can be found in Table A.7 within Appendix A. A total of 17 different features were proposed based on the Bradford Hill considerations and another 10 are proposed based on hierarchal structures. For some of the factors we considered numerous similar features that may be correlated as this is the first extensive study combining features based on the Bradford Hill considerations and it is not know which features may be more suitable. We investigated the importance of the features based on the average total decrease in Gini impurity gained by splitting the tree's node on a particular feature within the random forest (the importance function in the R randomForest library) [31].

The data used to generate the features includes the unique ID of the patient, the patient's age, the patient's gender, the dosage of the drug and a value indicating how noisy the data point may be due to how many other drugs the patient was prescribed in the month before and after the prescription. To calculate the features we extract certain records from the THIN database,

- $x^i_{K'}$ = (*patientID*,*age*,*gender*,*dosage*,*noise*) is a vector corresponding the details of the *k*th prescription of *$drug_i$* within the database.
- $y^i_{K'}$ = (*patientID*,*age*,*gender*,*dosage*,*noise*) is a vector corresponding the details of the *k*th prescription of *$drug_i$* within the database but where there has been no recording of the same drugcode within the past 13 months for the patient.

- $z^{i}_{k'\cdot}$ = (*patientID*,*age*,*gender*,*dosage*,*noise*) is a vector corresponding the details of the *k*th prescription of $drug_i$ within the database but where there has been no recording of any drug from the same drug family within the past 13 months for the patient.

The age is the patient's age when prescribed the drug, the gender is 1 if male and 0 otherwise and the dosage is the dosage of the prescription. The noise value corresponds to the number of other drugcodes that are prescribed for the same patient within 30 days before and 30 days after the drug prescription of interest. The instances when a patient experiences a medical event ($Readcode_j$) within a hazard period centred on the drug ($drug_i$) are,

- $x^{[u,v],i,j}_{k}$ = (*patientID*,*age*,*gender*,*dosage*, *noise*, *first*3, *first*4) is a vector corresponding the details of the *k*th time $Readcode_j$ is recorded between *u* and *v* days after the $drug_i$ is recorded within the database.

- $y^{[u,v],i,j}_{k}$ = (*patientID*,*age*,*gender*,*dosage*, *noise*, *first*3, *first*4) is a vector corresponding the details of the *k*th time $Readcode_j$ is recorded between *u* and *v* days after $drug_i$ is recorded for the first time in 13 months for the patient within the database.

- $z^{[u,v],i,j}_{k}$ = (*patientID*,*age*,*gender*,*dosage*, *noise*, *first*3, *first*4) is a vector corresponding the details of the *k*th time $Readcode_j$ is recorded between *u* and *v* days after the $drug_i$ is recorded and no drug from the same drug family is recorded for the patient within the previous 13 months within the database.

In the above vectors the first five elements correspond to the same details as the prescription vector but the binary element first3 is 1 iff there are no previous recordings of any *Readcode* □ $Readcode_j$ for the patient and first4 is 1 iff there are no previous recordings of any *Readcode* □ $Readcode_j$ for the patient.

We define $X^{i,\cdot} = \{x^{i}_{1'\cdot},...,x^{i}_{m'\cdot}\}$ to be the set of all vectors detailing each prescription of $drug_i$, so $|X^{i,\cdot}|$ is the total number of prescriptions of $drug_i$. $Y^{i,\cdot}$ and $Z^{i,\cdot}$ are similarly used to denote the set of vectors detailing each prescription of $drug_i$ such that the drug has not been prescribed within the previous 13 months or a similar drug has not been recorded within the previous 13 months respectively. We define $X^{[u,v],i,j} = \{x^{[u,v],i,j}_{1},...,x_m^{[u,v],i,j}\}$, so the cardinality of $X^{[u,v],i,j}$ (denoted $|X^{[u,v],i,j}|$) corresponds to the number of prescription of $drug_i$ that have the $Readcode_j$ recorded within their [*u*,*v*] hazard period. Similarly, $Y_{[u,v],i,j} = \{y_{[1u,v],i,j},...,y_{[mu,v],i,j}\}$ and $Z_{[u,v],i,j} = \{z_{1[u,v],i,j},...,z_{[mu,v],i,j}\}$.

The proposed features are,

(i) *Strength* - This factor represents the association strength, as generally a causal relationship is likely to have a large value of association, so a higher association means a higher likelihood of a causal relationship. A common way to calculate the association strength for drug safety (which could also be applied in general) is by investigating variation in the risk of experiencing a health outcome in patients exposed to a drug (or antecedent) and those unexposed.

The risk of $Readcode_j$ during a defined time period after $drug_i$ for a specific set of THIN records is simply the number of prescriptions of $drug_i$ in the set where $Readcode_j$ was recorded within a defined time period after $drug_i$ divided by the total number of prescriptions of $drug_i$ in the set. Similarly, the risk of $Readcode_j$ during a defined time period after any drug other than $drug_i$ for a specific set of THIN records is the number of prescriptions of any non-$drug_i$ drug in the set where $Readcode_j$ was recorded within the defined time period afterwards divided by the total number of prescriptions of any non-$drug_i$ drug in the set.

The risk difference is the risk of the health outcome in the group of patients exposed to the drug minus the risk of the health outcome in some control group of patients. The risk difference used in this study is the risk of $Readcode_j$ during the 1 to 30 day period after $drug_i$ minus the risk of the $Readcode_j$ during the 1 to 30 day period after any other drug. The 1 to 30 day period was chosen as we are interested in acutely occurring ADRs, but this period chould be adjusted to signal ADRs that take longer to occur. The formal calculations for the different set of THIN records X,Y and Z are,

$$
\begin{aligned}
Attr_{1i,j} &= (|X_{[1,30],i,j}|/|X_{i,.}|) - \left(\sum_{s,i}|X_{[1,30],s,j}| \Big/ \sum_{s,i}|X_{s,.}|\right)\\
Attr_{2i,j} &= (|Y_{[1,30],i,j}|/|Y_{i,.}|) - \left(\sum_{s,i}|Y_{[1,30],s,j}| \Big/ \sum_{s,i}|Y_{s,.}|\right) \qquad (1)\\
Attr_{3i,j} &= (|Z_{[1,30],i,j}|/|Z_{i,.}|) - \left(\sum_{s,i}|Z_{[1,30],s,j}| \Big/ \sum_{s,i}|Z_{s,.}|\right)
\end{aligned}
$$

(ii) *Specificity* - The specificity factor has been interpreted in many ways. The original interpretation is that it investigates whether the drug is observed to cause one or many medical events. There has been a debate regarding this consideration as some researchers believe it to be generally uninformative [32]. However, it has been argued that it can have a use for identifying causal relationships [33]. Some researchers consider the specificity to correspond to how specific the relationship is, for example is the association mainly found in a certain age and gender subpopulation or is the drug associated to a very specific medical event? One way to easily calculate whether the association tends to occur for a specific age range is to calculate the average age of the patients experiencing the medical event within the 1 to 30 day period after the drug and compare this to the average age of all the patients prescribed the drug,

$$Attr_4 = \left(m \times \sum_{k=1}^{n^{[1,30],i,j}} x_{k2}\right) / \left(n \times \sum_{k=1}^{m^{i,.}} x_{k2}\right)$$

$$Attr_5 = \left(m \times \sum_{k=1}^{n^{[1,30],i,j}} y_{k2}\right) / \left(n \times \sum_{k=1}^{m^{i,.}} y_{k2}\right) \quad (2)$$

$$Attr_6 = \left(m \times \sum_{k=1}^{n^{[1,30],i,j}} z_{k2}\right) / \left(n \times \sum_{k=1}^{m^{i,.}} z_{k2}\right)$$

Another specificity consideration is whether the association occurs more for one specific gender, an easy way to calculate a measure for this is to compare the fraction of patient that experience the medical event within the 1 to 30 day period after the drug who are male divided by the fraction of patients that are prescribed the drug who are male,

$$Attr_7 = \left(m \times \sum_{k=1}^{n^{[1,30],i,j}} x_{k3}\right) / \left(n \times \sum_{k=1}^{m^{i,.}} x_{k3}\right)$$

$$Attr_8 = \left(m \times \sum_{k=1}^{n^{[1,30],i,j}} y_{k3}\right) / \left(n \times \sum_{k=1}^{m^{i,.}} y_{k3}\right) \quad (3)$$

$$Attr_{9i,j} = \left(m \times \sum_{k=1}^{m} z_{k[13,30],i,j}\right) / \left(n \times \sum_{k=1} z_{ik,.3}\right)$$

and the final specificity feature is how specific the medical event is, this can be determined by the level of the Read code. We create a feature indicating whether the Read code is a level 5, level 4, level 3 or level 2 Read code,

if $Readcode_j$ is a level 5 Read code
if $Readcode_j$ is a level 4 Read code
if $Readcode_j$ is a level 3 Read code
if $Readcode_j$ is a level 2 Read code
otherwise (4)

(iii) *Temporality* - This factor investigates the direction of the relationship. If the relationship is causal then the drug must occur before the medical event. In [34], the authors used temporality features to train a classifier to discriminate between indicators and adverse events. This has also been seen in [35], where the authors used the temporality measures to identify off-target drugs. Similarly, for this factor we consider how often $Readcode_j$ occurs during the 1 to 30 day period after a prescription of $drug_i$ compared to the 1 to 30 day period before $drug_i$,

$$Attr_{11}^{i,j} = |X_{[1,30],i,j}| / |X_{[-30,-1],i,j}|$$

$$Attr_{12}^{i,j} = |Y_{[1,30],i,j}| / |Y_{[-30,-1],i,j}| \quad (5)$$

$$Attr_{13}^{i,j} = |Z_{[1,30],i,j}| / |Z_{[-30,-1],i,j}|$$

(iv) *Biological gradient* - In the context of detecting ADRs this relates to the dosage of the drug. It is generally the case, but not always [36], that there is a monotonic increasing relationship between the dosage of the drug and the probability of experiencing the ADR [37]. Due to this, Bradford Hill's criteria suggest that the dosage should be considered when determining causality, as medical events that occur more with high dosages are more likely to correspond to ADRs. A simple and efficient feature of the biological gradient criterion is to compare the average dosage given to patients who experienced the medical event within a month of taking the drug divided by the average dosage of the patients prescribed the drug,

$$Attr_{14}^{i,\cdot} = \left(m \times \sum_{k=1}^{n} x_{k4}\right) / \left(n \times \sum_{k=1}^{m_{i,j}} x_{k4}\right), \quad x \in X_{[1,30],i,j} \quad (6)$$

$$Attr_{15}^{i,\cdot} = \left(m \times \sum_{k=1}^{n} y_{k4}\right) / \left(n \times \sum_{k=1}^{m_{i,j}} y_{k4}\right), \quad y \in Y_{[1,30],i,j} \quad (7)$$

$$Attr_{16}^{i,\cdot} = \left(m \times \sum_{k=1}^{n} z_{k4}\right) / \left(n \times \sum_{k=1}^{m_{i,j}} z_{k4}\right), \quad z \in Z_{[1,30],i,j} \quad (8)$$

It could be possible to generate an improved feature of the biological gradient by investigating the correlation between the time to the health outcome and the dosage. Unfortunately this would be inefficient when analysing hundreds or thousands of drug and health outcome pairs, so the comparison of the averages were chosen instead.

(v) *Experimentation* - This has been openly interpreted. Some people believe this factor refers to using results of experiments, such as clinical trials, while others consider it to correspond to investigating the outcome from when a patient has a repeat of the antecedent event [36]. For the latter, in the context of ADRs, it is clear that when a patient experiences an ADR it will occur every time after the drug is ingested (under the same conditions) but should not be present when the drug stops being ingested. Therefore, if the drug is prescribed two or more distinct times (break of 12 months or more between prescriptions) for a patient and the medical event always follows but stops when the drug stops, then this would be strong evidence to suggest a causal relationship. Therefore, we calculate the number of people who experience $Readcode_j$ within a 1 to 30 day period after $drug_i$ for two or more distinct prescriptions and never during the 1 to 30 day period before divided by the number of patients who have two or more distinct prescriptions,

$$|\{y_{[1_k 1,30],i,j} \,|\, y_{[1_k 1,30],i,j} \in \bigcup y_{[1_s 1,30],i,j}\} \cap (\bigcup y_{[s-1\,30,-1],i,j})^c\}|$$

$$Attr17_{i,j} = \frac{|\{y_{k,1}^{i,.} \in S_s\}|}{|y_{k1} \in S_{s,k}\, y_{s1}|} \quad (9)$$

(vi) *Hierarchal specific* - We also propose features for dealing with the hierarchical medical event coding structures found within THIN (and also in many other medical databases). The first feature is related to the noise caused by patients taking multiple prescriptions. If a patient is taking more than one drug then the medical event may be caused by the other drug, so the more drugs a patient is taking the higher the risk of confounding due to other drugs. The noise value in the prescription vector tells us how many others drugs the patient has taken around the prescription, so we calculate the average number of other drugs recorded within the 1 to 30 day period before or after $drug_i$ for the patients experiencing $Readcode_j$ divided by the average over all prescriptions of $drug_i$,

$$Attr_{18} = \left(m \times \sum_{k=1}^{n} x_{k5}^{[1,30],i,j}\right) \Big/ \left(n \times \sum_{k=1}^{m_{i,j}} x_{k5}^{i,.}\right) \quad (10)$$

$$Attr_{19} = \left(m \times \sum_{k=1}^{n} y_{k5}^{i,.}\right) \Big/ \left(n \times \sum_{k=1}^{m_{i,j}} y_{k5}^{[1,30],i,j}\right) \quad (11)$$

$$Attr_{20} = \left(m \times \sum_{k=1}^{n} z_{k5}^{i,.}\right) \Big/ \left(n \times \sum_{k=1}^{m_{i,j}} z_{k5}^{[1,30],i,j}\right) \quad (12)$$

To deal with the hierarchical structures we also generate features that can give insight into when the association might correspond to a medical event that has previously been recorded but as a more general Read code. Firstly we calculate the number of times the level 3 version of $Readcode_j$ is recorded for the first time ever within a 1 to 30 day period after $drug_i$ divided by the number of times the level 4 version of $Readcode_j$ is recorded for the first time within the 1 to 30 day period after $drug_i$,

$$Attr21_{i,j} = \sum_{p=1}^{n} x_{[1p6,30],i,j} \Big/ \sum_{p=1}^{n} x_{[1p7,30],i,j} \quad (13)$$

The final two features consider the temporality measure but when considering more general versions of the Read code. These features calculate similar values to those used in the temporality section but when reducing all the Read codes to their level 4 versions or level 3 versions respectively,

$$Attr_{22}^{i,j} = \left| \bigcup_{\{k|k \in j^4\}} X_{[1,30],i,k} \right| / \left| \bigcup_{\{k|k \in j^4\}} X_{[-30,-1],i,k} \right| \quad (14)$$

$$Attr_{23}^{i,j} = \left| \bigcup_{\{k|k \in j^3\}} X_{[1,30],i,k} \right| / \left| \bigcup_{\{k|k \in j^3\}} X_{[-30,-1],i,k} \right| \quad (15)$$

$$Attr_{24}^{i,j} = \left| \bigcup_{\{k|k \in j^4\}} Y_{[1,30],i,k} \right| / \left| \bigcup_{\{k|k \in j^4\}} Y_{[-30,-1],i,k} \right| \quad (16)$$

$$Attr_{25}^{i,j} = \left| \bigcup_{\{k|k \in j^3\}} Y_{[1,30],i,k} \right| / \left| \bigcup_{\{k|k \in j^3\}} Y_{[-30,-1],i,k} \right| \quad (17)$$

$$Attr_{26}^{i,j} = \left| \bigcup_{\{k|k \in j^4\}} Z_{[1,30],i,k} \right| / \left| \bigcup_{\{k|k \in j^4\}} Z_{[-30,-1],i,k} \right| \quad (18)$$

$$Attr_{27}^{i,j} = \left| \bigcup_{\{k|k \in j^3\}} Z_{[1,30],i,k} \right| / \left| \bigcup_{\{k|k \in j^3\}} Z_{[-30,-1],i,k} \right| \quad (19)$$

After calculating the features for each $drug_i$-$Readcode_j$ pair we have the corresponding feature vector $Attr^{i,j} \in \mathbb{R}^{27}$, $Attr^{i,j} = (Attr_1^{i,j}, Attr_2^{i,j}, ..., Attr_{27}^{i,j})$. The Bradford Hill consideration features are $Attr^{i,j} \in \mathbb{R}^{17}$, $Attr^{i,j} = (Attr_1^{i,j}, Attr_2^{i,j}, ..., Attr_{17}^{i,j})$. The hierarchal knowledge features are $Attr^{i,j} \in \mathbb{R}^{10}$, $Attr^{i,j} = (Attr_{18}^{i,j}, Attr_{26}^{i,j}, ..., Attr_{27}^{i,j})$. The association strength features are $Attr^{i,j} \in \mathbb{R}^{3}$, $Attr^{i,j} = (Attr_1^{i,j}, Attr_2^{i,j}, Attr_3^{i,j})$.

The majority of the features described above investigate the 30 day period after the drug. This was due to our problem focussing on signalling ADRs that occur within 30 days of the drug prescription. For signalling ADRs that occur over a larger time period, the proposed features can simply be modified by adjusting the length of the time period observed after the potential cause was recorded from [1, 30] to any time period of any length ([*a*,*b*]).

Table 2: The feature subsets used for comparison to determine the usefulness of including more Bradford Hill considerations and features related to the hierarchal structure of the database..

| Method | Number of features | Feature set name | Features included |
|---|---|---|---|
| SADR | 27 | All features | $Attr_1 - Attr_{27}$ |
| Comparison 1 | 17 | Bradford Hill | $Attr_1 - Attr_{17}$ |
| Comparison 2 | 10 | Hierarchal | $Attr_{18} - Attr_{27}$ |
| Comparison 3 | 3 | Association strength | $Attr_1 - Attr_3$ |

### *2.2.3. Step 3: Train the classifier*

After engineering the features for each of the drug-Read code pairs from the OMOP NSA reference set, where the Read code was recorded at least 3 times within 30 days after the drug in THIN, we trained a random forest classifier. To fairly evaluate the SADR framework, we decided to measure the trained classifier's performance on each drug separately. This was accomplished by creating 9 training/testing sets. Each training set contained all the data-points for 8 of the drugs and its corresponding testing set contained all the data points for the remaining drug. If we trained the classifier on all the labelled data then it would not be possible to validate that the classifier would work on new data. However, leaving out the labelled data for one drug during training and then applying the trained classifier on the left out drugs data to compare the prediction and truth is effectively mimicking the situation of applying the classifier to new data.

The software used for the classification was R and the 'caret' [38] library. A parameter grid search was applied to find the optimal number of parameters to use for each decision tree in the forest (the mtry parameter) and 10-fold cross validation is implemented to reduce over fitting.

## *2.3. Evaluation Method*

To investigate the importance of various proposed features, we investigate the performance of the SADR framework (training a random forest using both the Bradford Hill consideration and hierarchal-based features) with a random forest trained using only the Bradford Hill consideration based features or only the hierarchal features or only the association strength features. Table 2 lists the various feature subsets investigated for comparison and details the included features.

The performance measures we used were the same as in the previous study [39] using the NSA reference set to enable a comparison. Using the classifiers prediction and the ground truth, the number of true positives (TP) is the number of drug-HOIs where the classifier predicts the drug-HOIs are ADRs and the ground truth is that the drug-HOIs are ADRs. The number of false positives (FP) is the number of drug-HOIs where the classifier predicts the drug-HOIs are ADRs and the ground truth is that the drug-HOIs are non-ADRs. The number of false negatives (FN) is the number of drug-HOIs where the classifier predicts the drug-HOIs are non-ADRs and the ground truth is that the drug-HOIs are ADRs. Finally, the number of true negatives (TN) is the number of drug-HOIs where the classifier predicts the drug-HOIs are non-ADRs and the ground truth is that the drug-HOIs are non-ADRs.

The area under the receiver operating characteristic (ROC) curve (AUC) considers the trade off between the sensitivity of the classification and the specificity. ROC curves are drawn by plotting the sensitivity (TP/(TP+FN)) against one minus the specificity (TN/(TN+FP)) at different threshold stringencies. The AUC of the ROC plot gives an indication of how well the classifier performs. To compare the classifiers the DeLong's test at a 5% significance level [40] is applied to the AUCs. The DeLong's test is a nonparametric inference that is applied to determine whether the AUC between two paired ROC curves is significantly different. The average precision (AP), precision at cutoff 10 ($P_{10}$), false positive rate (FPR=FP/(FP+TN)) and $Recall_5$ (=TP/(TP+FN)) at a FPR of 5% are also calculated.

## 3. Results

### *3.1. Evaluation of the SADR framework*

The results are presented in Table 3. The first three columns detail the data used for testing, stating which drug was excluded from the training data and how many drug-Read code pairs corresponding to ADRs and non-ADRs (nADR) were in the testing set. The next four columns detail the training set, the optimal mtry found by tuning the random forest on the training set using 10-fold cross validation using the AUC as the performance measure and the classifier's AUC (plus standard deviation) obtained by the cross-validation. The final five columns present the various measures of performance of the trained random forest on the test set (effectively its performance on new data). Figures 2-4 show the ROC plots of the SADR framework.

The combination of Bradford Hill and hierarchal features used as inputs into a random forest for signalling ADRs resulted in an average AUC of 0.862 and MAP of 0.640 across all nine drug test sets. The top 10 drug-Read code pairs ranked by the classifier's confidence of the pair belonging to the ADR class were all true ADRs for 6 out of the 9 drug test sets. The FPR ranged between 0.000 to 0.026 at the classifier's natural threshold and at a 5% FPR the average recall was 0.583.

*3.2. Comparison of the SADR framework against random forests with other feature sets*

The SADR framework (random forest using all the Bradford Hill and hierarchal features) had higher minimum, maximum and mean AUCs, 0.792, 0.940 and 0.862 respectively, across the 9 test sets compared to the random forests using only the Bradford Hill features (comparison 1), only the hierarchal features (comparison 2) or only the association strength features (comparison 3), see Table 4. The p-values comparing the AUC of the SADR framework with the three comparisons for each drug test set are displayed in Table 5. The SADR framework was significantly better than the random forest using association strength features for all drugs except the Tricyclic antidepressants, although this p-value approached significance (0.0584). The SADR framework was significantly better than the random forest using the hierarchal features for the majority of drugs (6 out of the 9 drugs). However, the SADR framework was only significantly better than the random forest using the Bradford Hill features for 4 out of the 9 drugs.

Table 3: Performance of SADR framework when trained on 8 of the drugs and tested on the remaining drug. The mtry is the value corresponding to the optimal training model's mtry returned by the grid search when 10-fold cross validation is implemented.

| Test Set | | | Training Set & Performance | | | | Test Set Results | | | | |
|---|---|---|---|---|---|---|---|---|---|---|---|
| Drug | ADR | nADR | ADR | nADR | mtry | AUC (SD) | AUC | AP | $P_{10}$ | FPR | $Recall_5$ |
| OMOP Benzodiazepines | 36 | 342 | 369 | 3502 | 5 | 0.871(0.035) | 0.940 | 0.730 | 1.000 | 0.003 | 0.611 |
| OMOP Antiepileptics | 41 | 403 | 364 | 3441 | 5 | 0.879(0.045) | 0.792 | 0.442 | 0.700 | 0.007 | 0.439 |
| OMOP ACE Inhibitor | 76 | 491 | 329 | 3353 | 5 | 0.877(0.030) | 0.875 | 0.711 | 1.000 | 0.002 | 0.592 |
| OMOP Bisphosphonates | 44 | 258 | 361 | 3586 | 5 | 0.870(0.039) | 0.795 | 0.550 | 0.800 | 0.016 | 0.477 |
| OMOP Beta blockers | 38 | 492 | 367 | 3352 | 5 | 0.867(0.045) | 0.922 | 0.766 | 1.000 | 0.004 | 0.711 |
| OMOP Warfarin | 33 | 313 | 372 | 3531 | 10 | 0.881(0.036) | 0.863 | 0.704 | 1.000 | 0.000 | 0.606 |
| OMOP Typical antipsychotics | 40 | 301 | 365 | 3543 | 5 | 0.874(0.039) | 0.883 | 0.728 | 1.000 | 0.007 | 0.675 |
| OMOP Tricyclic antidepressants | 54 | 425 | 351 | 3419 | 5 | 0.883(0.039) | 0.828 | 0.598 | 1.000 | 0.005 | 0.574 |
| OMOP Antibiotics | 43 | 819 | 362 | 3025 | 15 | 0.879(0.037) | 0.862 | 0.533 | 0.800 | 0.026 | 0.558 |
| Overall Average | - | - | - | - | - | - | 0.862 | 0.640 | 0.922 | 0.008 | 0.583 |

The ADR and nADR columns indicate the number of drug-outcome pairs labelled as ADRs or non-ADRs in the test/training sets.

*3.3. Feature Importance*

The importance of the features is presented in Table 6. Interestingly the feature chosen as the most important is the experimentation feature (68.7 importance), followed by the association strength (44.6-47.9 importance) and the hierarchal features based on noise or the ratio to how often the Read code level 3 version occurs for the first time after the drug compared to the level 4 version. The highest of the specificity features is the average age of the patients prescribed any drug that experience the Read code divided by the average of the patients prescribed the drug. In

Table 4: Summary of the performances of the SADR framework compared with random forests trained on various subsets of the features proposed in Section 2.2.2 across the 9 drug test sets.

| Method | Feature Set | Min AUC | Max AUC | Average AUC |
|---|---|---|---|---|
| SADR | Bradford Hill & Hierarchal | 0.792 | 0.940 | 0.862 |
| Comparison 1 | Bradford Hill | 0.737 | 0.900 | 0.830 |
| Comparison 2 | Hierarchal | 0.721 | 0.869 | 0.787 |
| Comparison 3 | Strength Ensemble | 0.637 | 0.797 | 0.742 |

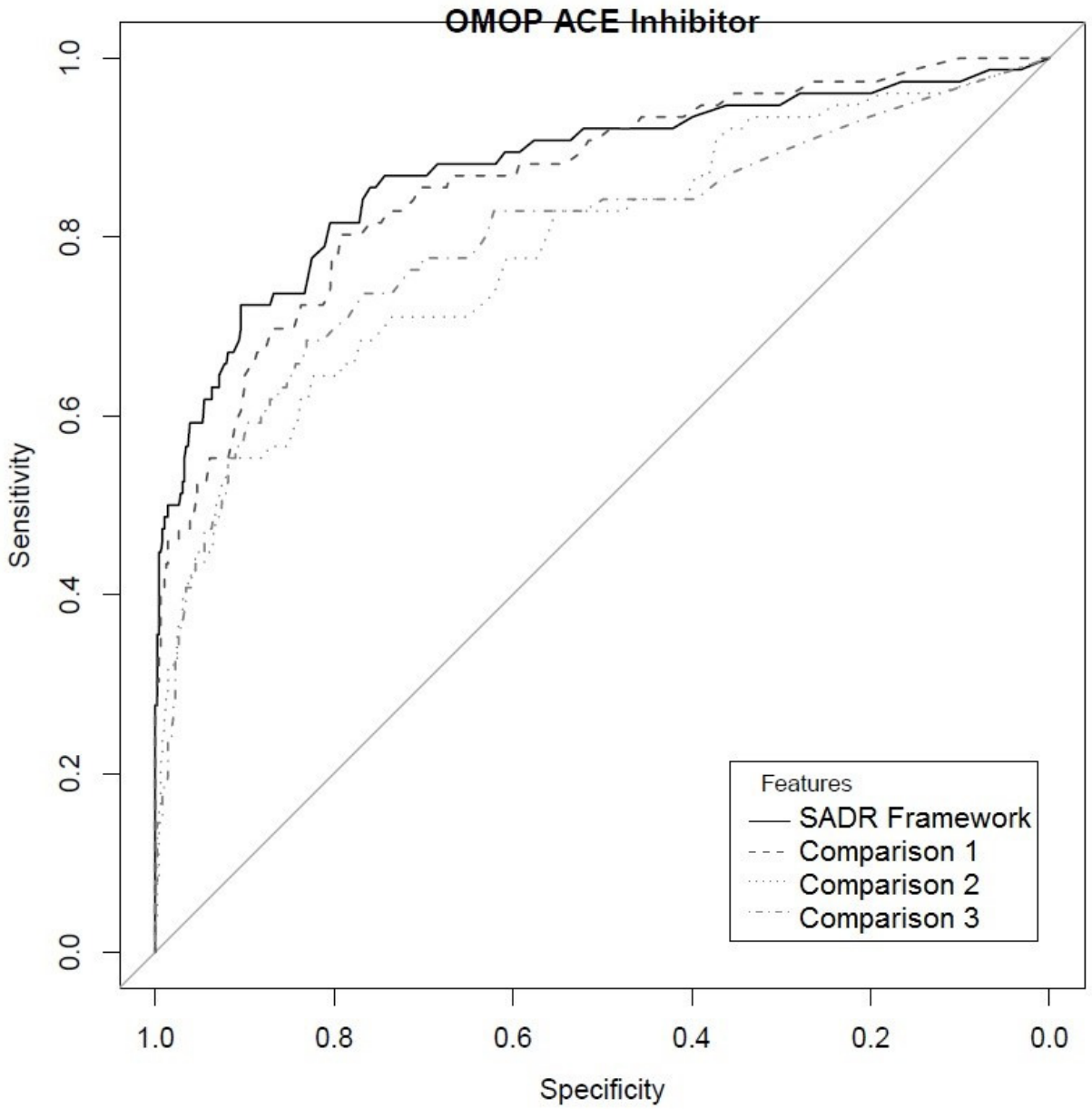

(a) Ace Inhibitors

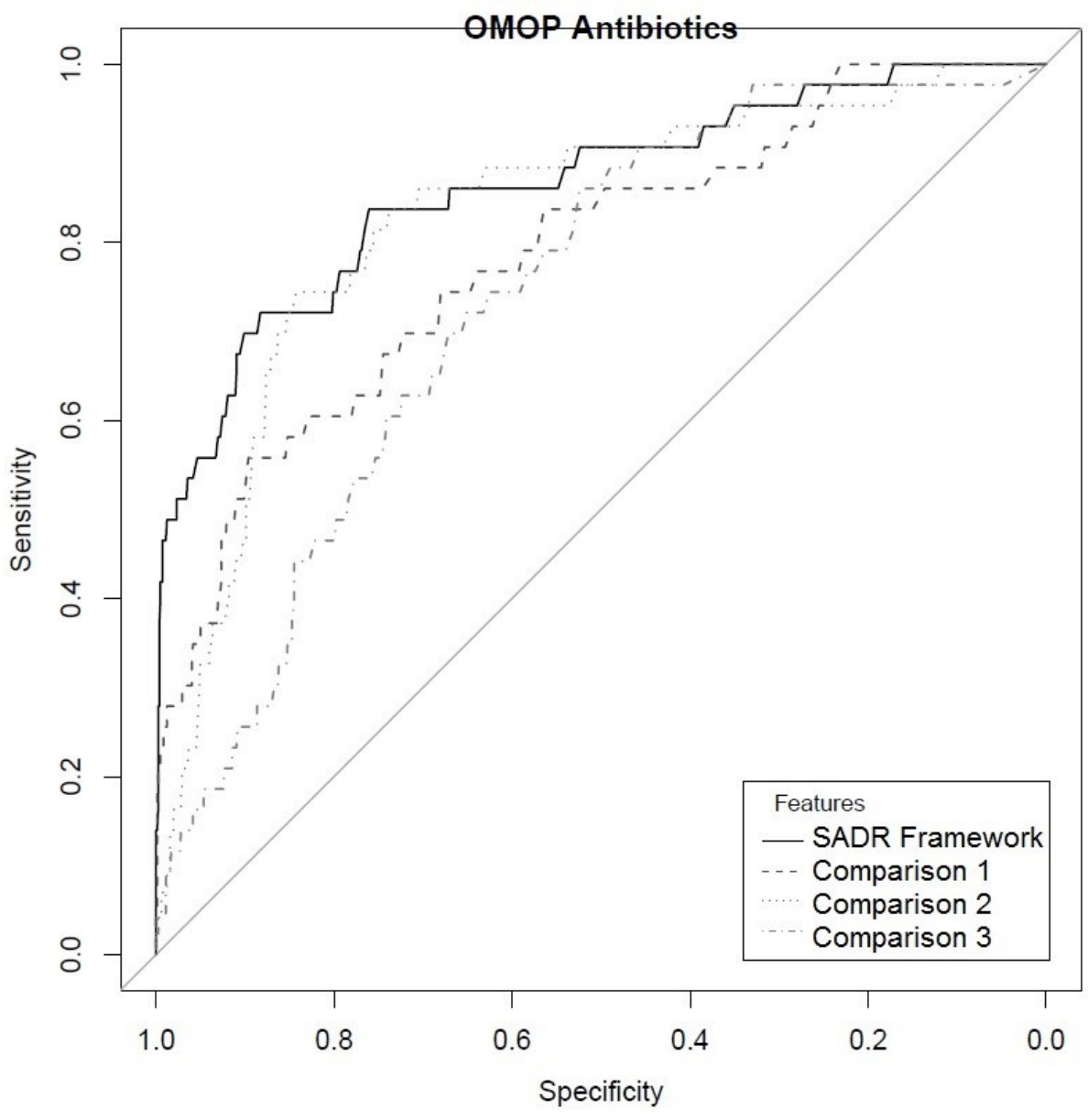

(b) Antibiotics

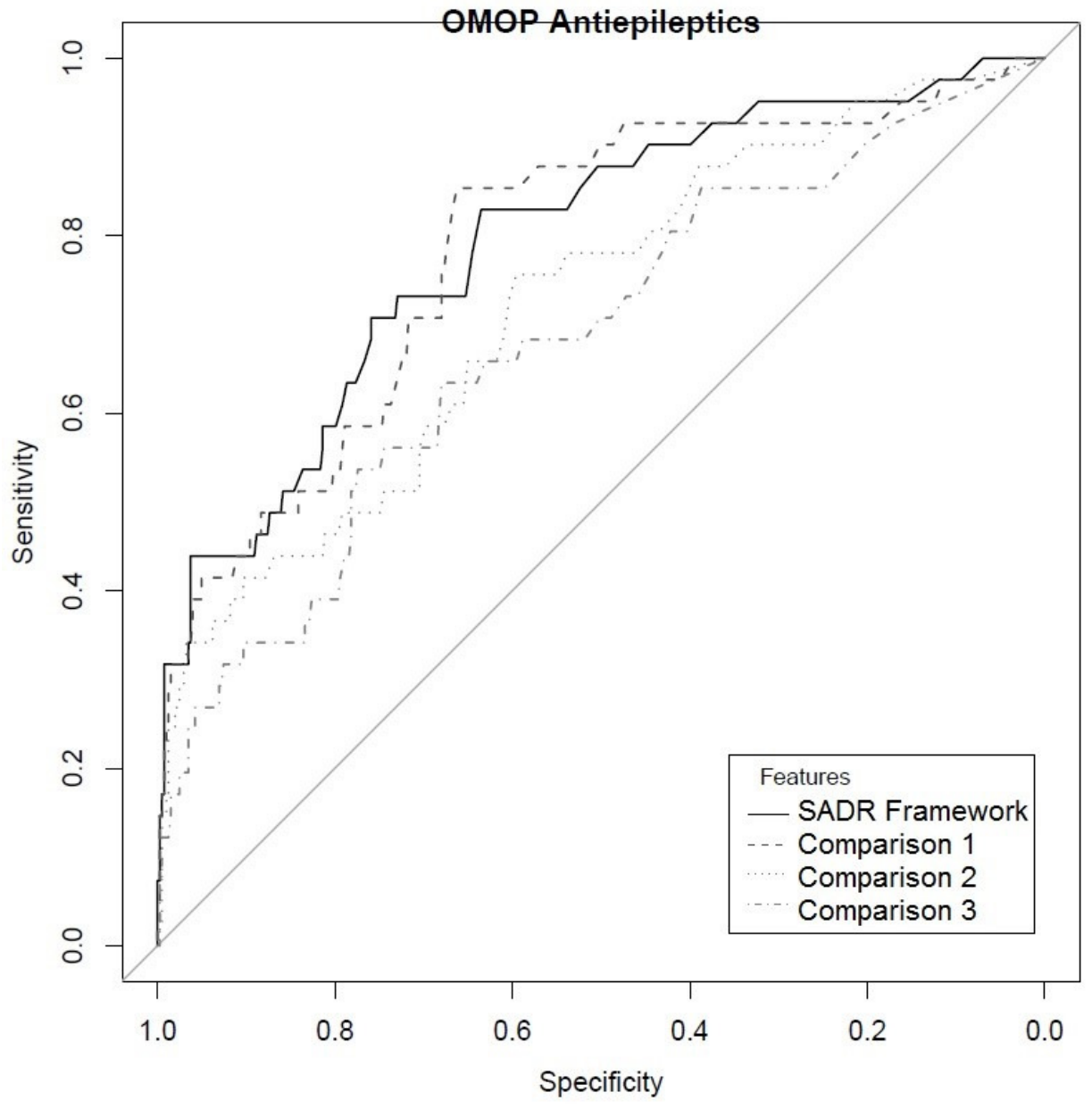

(c) Antiepileptics

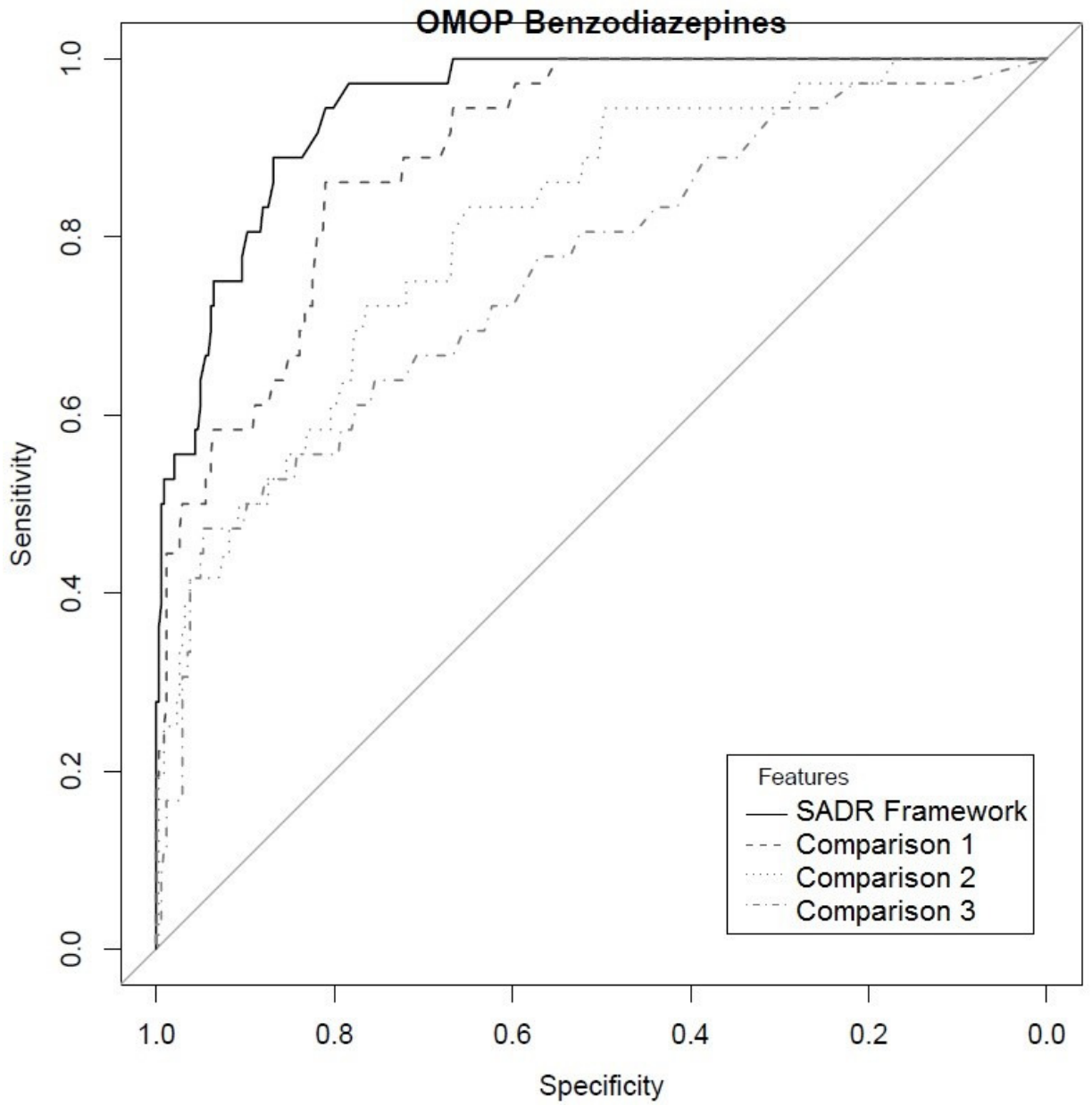

(d) Benzodiazepines

Figure 2: Reciever operating characteristic curve plots for the different drug test sets. Comparison between SADR framework (all features), Comparison 1 (Bradford Hill features only), Comparison 2 (hierarchal features only) and

Comparison 3 (Association strength features only).

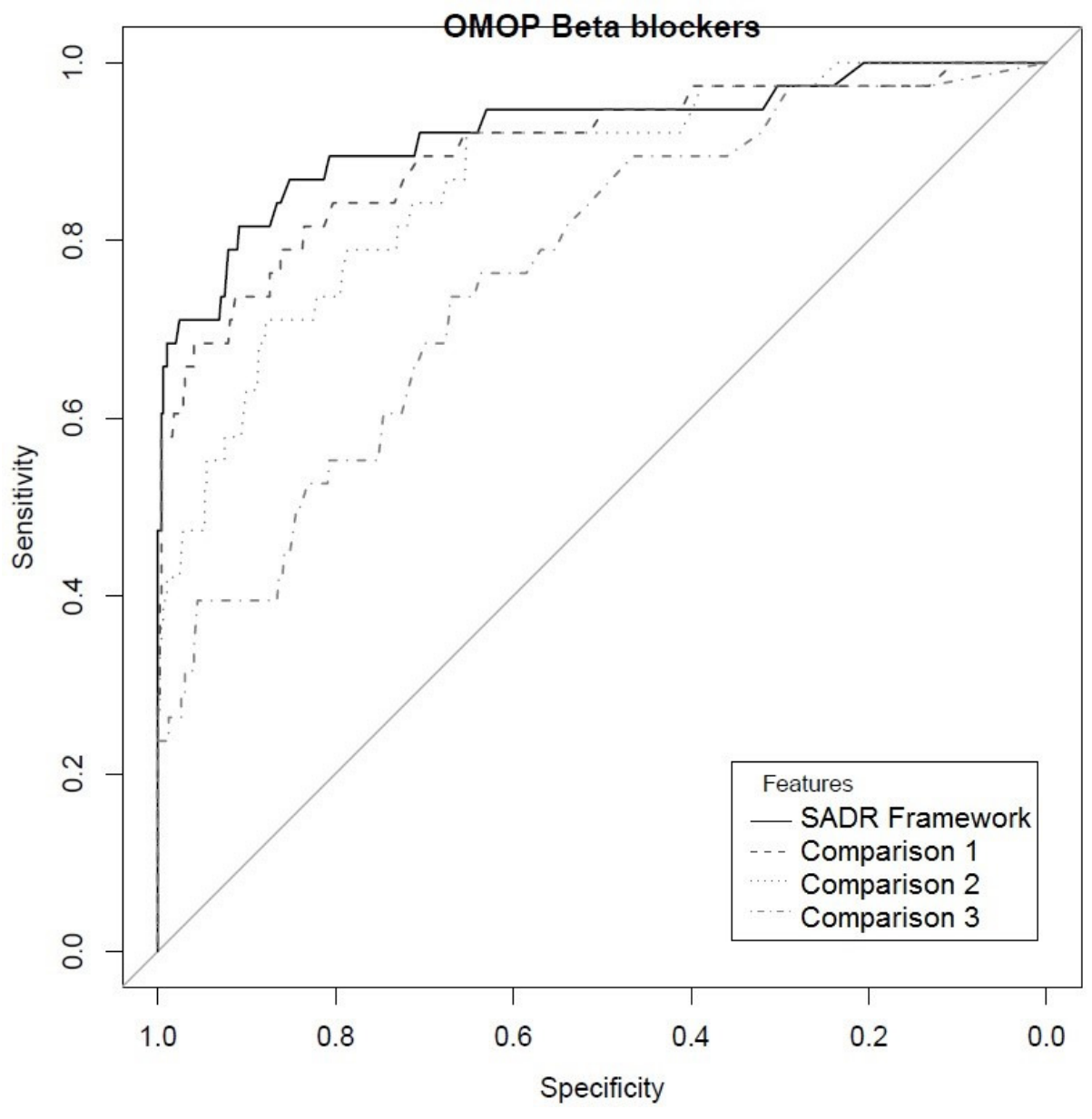

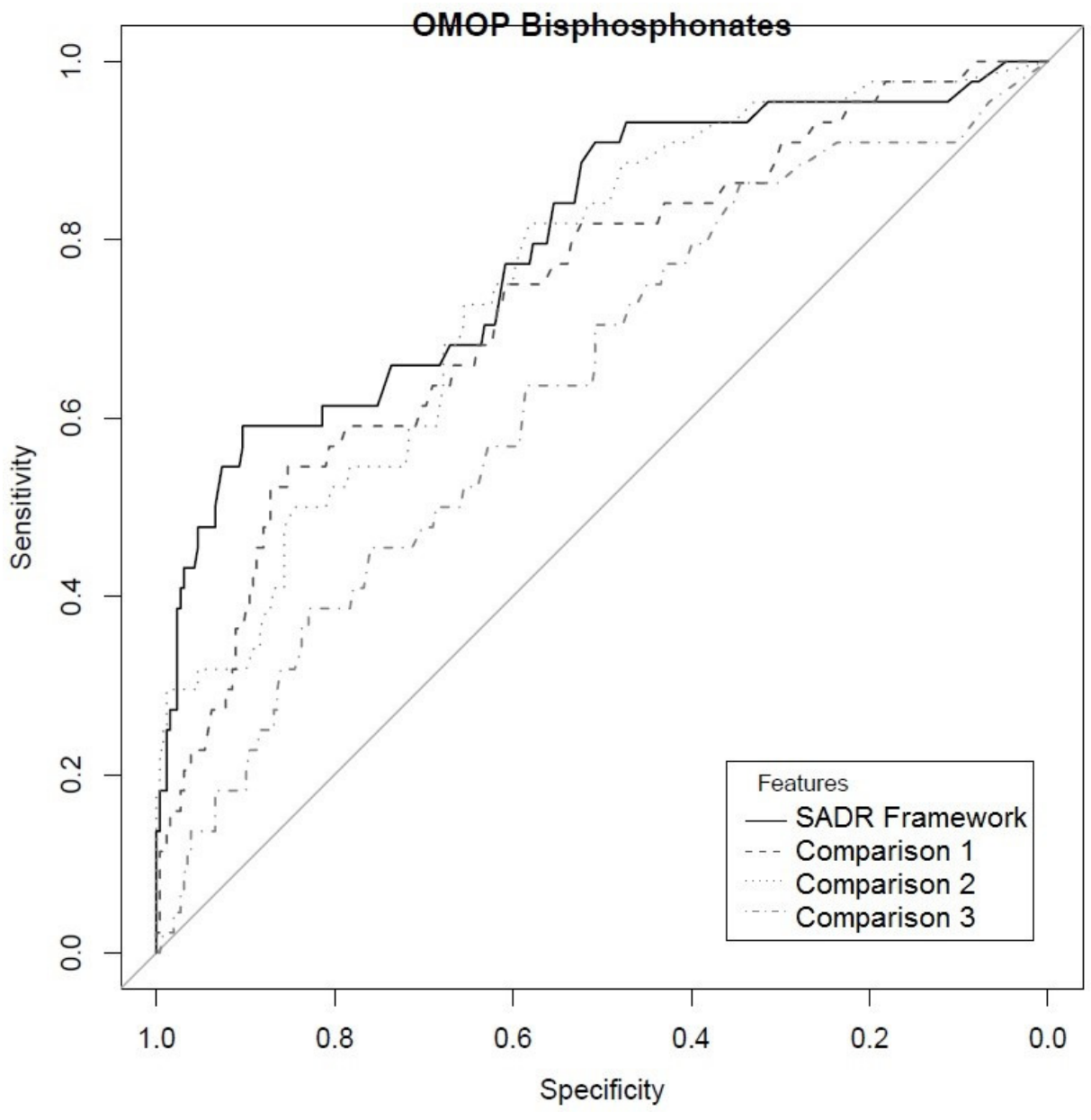

(a) Betablockers (b) Bisphosphonates

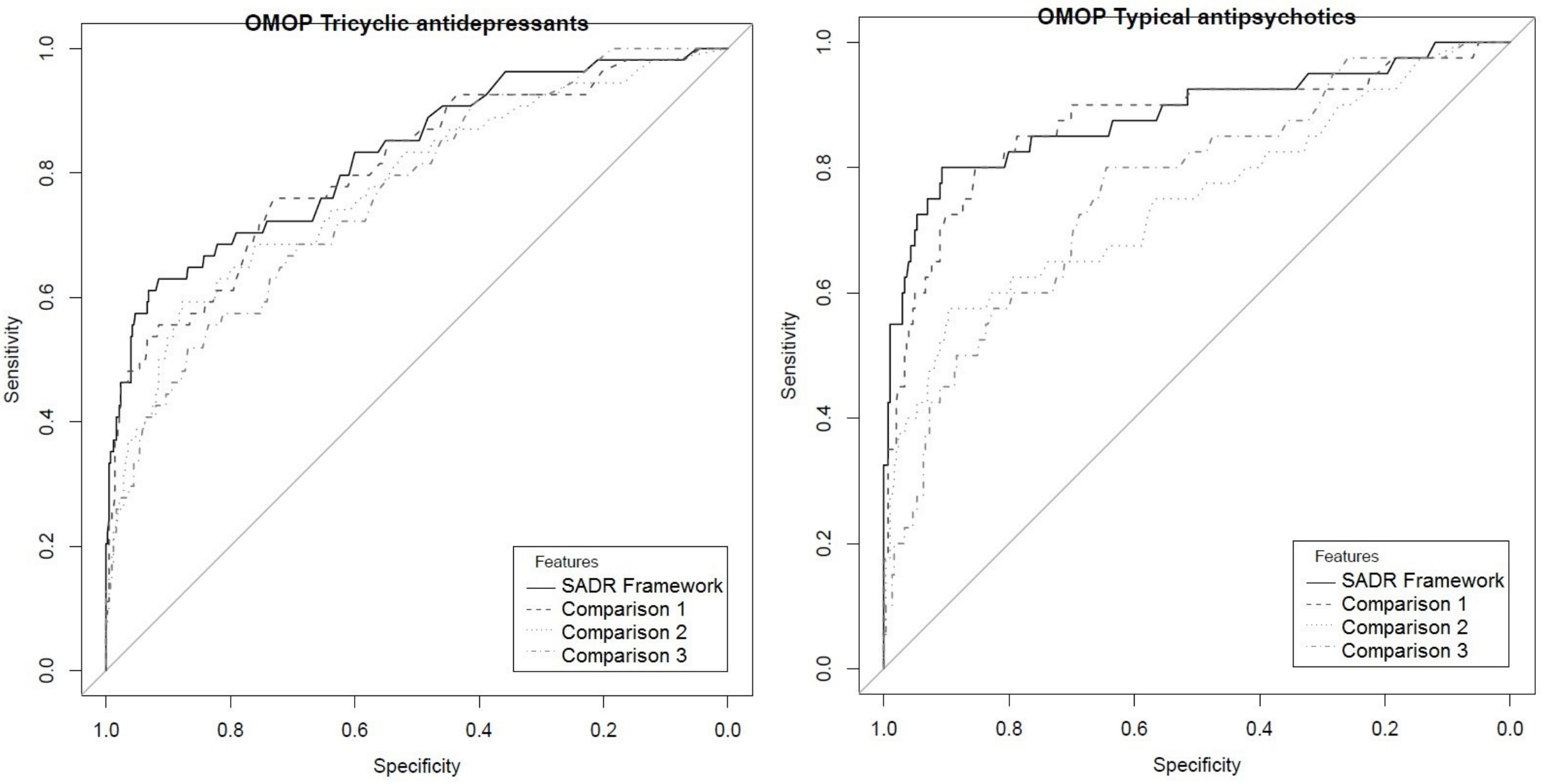

(c) Tricyclic antidepressants (d) Typical antipsychotics

Figure 3: Reciever operating characteristic curve plots for the different drug test sets. Comparison between SADR framework (all features), Comparison 1 (Bradford Hill features only), Comparison 2 (hierarchal features only) and

Comparison 3 (Association strength features only).

Table 5: The p-values from the DeLong's bootstrap test to compare the AUC for the paired ROC curves between the SADR framework and each comparison.

| Test Set | Comparison 1(Bradford Hill) | Comparison 2 (Hierarchal) | Comparison 3 (Strength Ensemble) |
|---|---|---|---|
| OMOP Benzodiazepines | 0.0046 | 0.0000 | 0.0000 |
| OMOP Antiepileptics | 0.8309 | 0.0445 | 0.0025 |
| OMOP ACE Inhibitor | 0.2307 | 0.0000 | 0.0116 |
| OMOP Bisphosphonates | 0.0010 | 0.1410 | 0.0000 |
| OMOP Beta blockers | 0.1033 | 0.0095 | 0.0000 |
| OMOP Warfarin | 0.0045 | 0.0115 | 0.0129 |
| OMOP Typical antipsychotics | 0.4236 | 0.0019 | 0.0011 |
| OMOP Tricyclic antidepressants | 0.3258 | 0.0755 | 0.0584 |
| OMOP Antibiotics | 0.0028 | 0.1974 | 0.0004 |

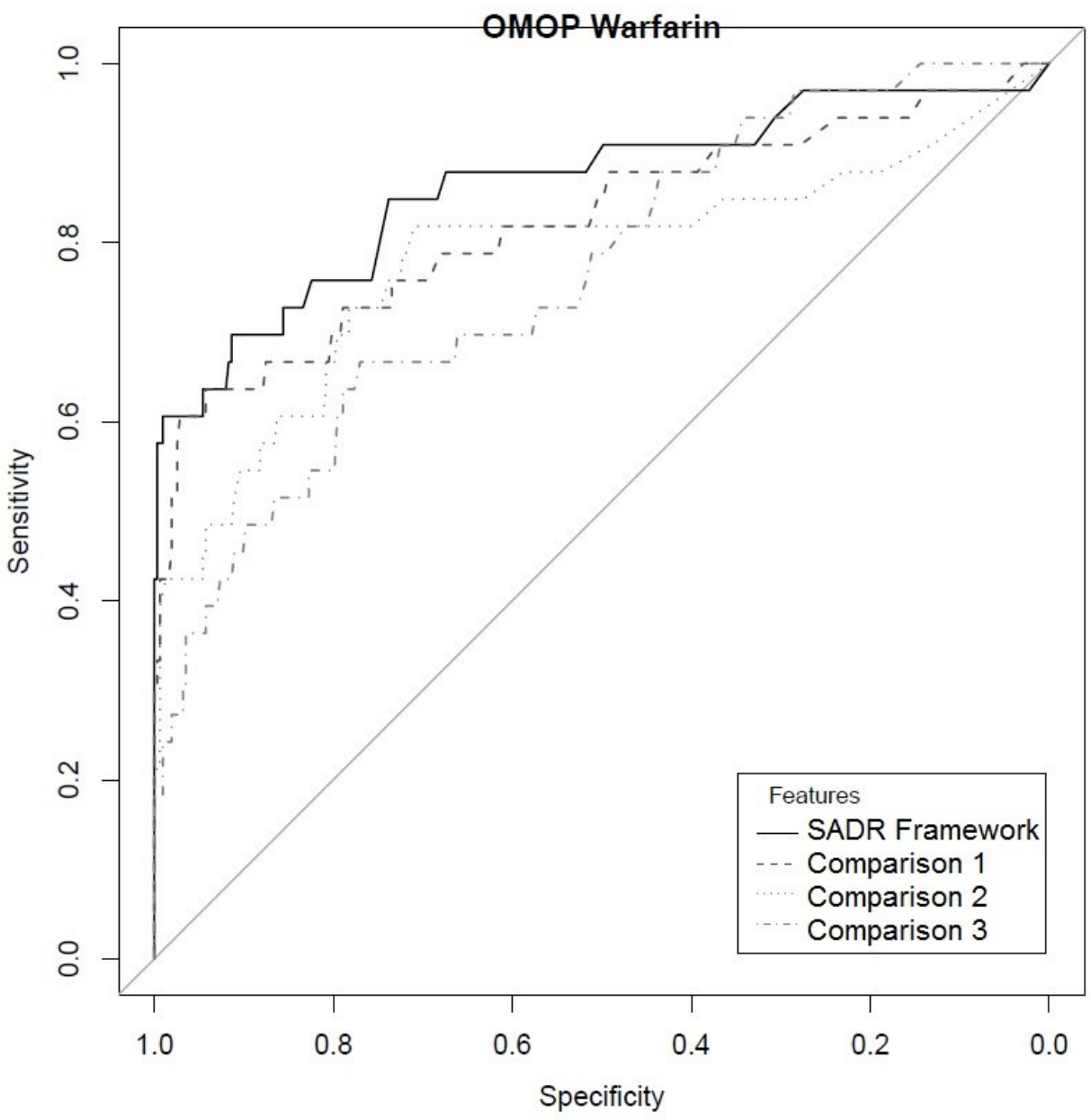

Figure 4: Reciever operating characteristic curve plots for the Warfarin test set.

general, the features that considered all the drug prescriptions tended to have a higher average gini decrease value.

The hierarchal features investigating how many time the level 3 version of the Read code occurs 30 days after the drug compared to 30 days before had the lowest mean gini decrease, ranging between 3.6 to 6. Apart from these features, all the other features seem to be useful for the model as they had average gini decrease values greater than 10.

## 4. Discussion

The results show that the SADR framework (a random forest using feature engineering based on the Bradford Hill considerations and hierarchal clinical codings) leads to the optimal classifier for discriminating between causality and association. This suggests that using the proposed features can enable supervised learning to be applied successfully to the problem of signalling ADRs in longitudinal observational data. The SADR framework obtained an average AUC of 0.862, a MAP of 0.640, an average FPR of 0.008 and an average recall at 5% FPR of 0.583 on the OMOP NSA reference set. In previous studies the Highthroughput Screening by Indiana University obtained the highest AUC of 0.734 [39], with MAP, FPR and recall at 5% FPR scores of 0.141, 0.266 and 0.367 respectively. This shows that a supervised approach made possible by the feature engineering can lead to improved ADR signalling. The supervised approach achieved a much lower false positive rate meaning it is extremely unlikely to signal non-ADRs.

Although the supervised approached outperformed the unsupervised methods, it may be argued that the additional complexity of the supervised approach makes it less suitable. However, we purposely suggested simple features that are quick to calculate and do not require tuning (e.g., once you know what time period of interest to investigate [*a*,*b*], there are no parameters to tune when calculating the features). Conversely, the existing unsupervised methods are generally more complex and often have parameters that need to be tuned for new datasets and this requires applying the methods numerous times. It is the data extraction and calculation of association strength (such as risk) from the big longitudinal observational data is often the time consuming aspect of both the unsupervised and supervised methods. Due to the simplicity of our proposed features, the framework combining the proposed feature engineering and classification is actually likely to be quicker that the existing unsupervised methods. In addition, the simplicity of the features also means that they could be calculated using distributed computing tool such as Hadoop [41]. In theory, this would make the feature extraction scalability quasi-linear and could make the framework suitable for many terabytes of data. Another advantage of the simplicity of the features is that it enables them to the updated efficiently with the addition of new data. For example, if we stored the number of patients prescribed the drug who experience the Read code within 30 days ($n_1$) as well as the average age of the patients prescribed the drug who experience the Read code within 30 days $avAge_1$, when new data are added we can simply calculate how many new occurrences of the Read code occur within 30 days of the drug $n_2$ and the average ages of these $avAge_2$ to quickly update the new values $n = n_1 + n_2$, $avAge = (n_1 \times avAge_1 + n_2 \times avAge_2)/(n_1 + n_2)$. This means we only have to extract the features on the big data once, and then we can update the features with the addition of small amounts of new data.

Table 6: The importance of the features within the random forest based on the total decrease in Gini impurity

| Category | Set | Att | Gini decrease |
|---|---|---|---|
| Experimentation | X | $Attr_{17}$ | 68.7 |
| Association | Z | $Attr_{3}$ | 47.9 |
| Association | X | $Attr_{1}$ | 46.7 |
| Association | Y | $Attr_{2}$ | 44.6 |
| Heirarchal | X | $Attr_{18}$ | 41.7 |
| Heirarchal | X | $Attr_{21}$ | 33.2 |
| Specificity | X | $Attr_{4}$ | 29.6 |
| Heirarchal | Y | $Attr_{19}$ | 28.5 |
| Heirarchal | Z | $Attr_{20}$ | 28.4 |
| Specificity | X | $Attr_{7}$ | 25.8 |
| Specificity | Y | $Attr_{5}$ | 25.0 |
| Specificity | Z | $Attr_{6}$ | 23.0 |
| Heirarchal | X | $Attr_{22}$ | 21.4 |
| Specificity | Z | $Attr_{9}$ | 21.2 |
| Specificity | Y | $Attr_{8}$ | 20.5 |
| Temporality | X | $Attr_{11}$ | 20.4 |
| Biological Gradient | X | $Attr_{14}$ | 19.7 |
| Heirarchal | Z | $Attr_{26}$ | 16.9 |
| Biological Gradient | Y | $Attr_{15}$ | 16.6 |
| Heirarchal | Y | $Attr_{24}$ | 16.5 |
| Biological Gradient | Z | $Attr_{16}$ | 16.4 |
| Temperality | Y | $Attr_{12}$ | 14.1 |
| Temporality | Z | $Attr_{13}$ | 13.2 |
| Specificity | - | $Attr_{10}$ | 11.0 |
| Heirarchal | Z | $Attr_{27}$ | 6.0 |
| Heirarchal | Y | $Attr_{25}$ | 5.0 |
| Heirarchal | X | $Attr_{23}$ | 3.6 |

Interestingly, the results showed that the experimentation feature seemed to be good at discriminating between causal and non-causal relationships. It is widely accepted that randomised controlled trials, a form of experimentation, are the best way to identify causality. The results of this paper suggest there may be ways to use longitudinal observational data to perform a weak form of experimentation that is still informative. The hierarchal feature corresponding to how many other drugs a patient has around the time of prescription could actually be interpreted as corresponding to the original specificity definition. If a patient is only taking one drug, then the relationship between the drug and health outcome is more specific. This was an influential feature in the random forest, which suggests the original specificity definition may have merit, although many have argued against its use.

In this paper we did not consider any coherence, plausibility or consistency features. The consistency features could be generated by using alternative data, such as spontaneous reporting system databases, and calculating the association strength for the drug and medical event pair within that data. The plausibility and coherence features may be possible by using the chemical structure data, as this may indicate chemical structures that are associated to an ADR or by adding a human feedback loop where a medical expert can identify incorrectly labelled pairs. Another possibility to engineer features for plausibility would be to implement a crowd sourcing algorithm that can extract suspected ADRs from online forums or medical literature.

It is worth noting that the Bradford Hill considerations are known to be limited and are consider only as a guide for inferring causality [42]. The Bradford Hill considerations are not definitive criteria for causality and any limitation in discriminating between causal and non-causal relationships using these considerations will also likely be a limitation of the SADR framework. However, machine learning techniques such as random forest can identify complex patterns, ones that are unlikely to be identified by humans, which may reduce the limitations of the considerations compared to when they are implemented manually.

One final comment, is that the performance of supervise learning techniques will generally improve as the number of labelled data to train on increases. Therefore, the proposed framework is expected to improve over time as more labelled data becomes available.

## 5. Conclusion

In this paper we proposed a novel supervised ADR signalling framework (SADR) utilising Bradford Hill's causality considerations to enable the implementation of a classifier that can accurately signal ADRs in big longitudinal observational databases that suffer from confounding and have hierarchal clinical code structures.

The framework trains a random forest to discriminate between ADRs and non-ADRs using suitable features based on the Bradford Hill causality considerations (many of which have been previously used to signal ADRs but never combined). The trained random forest performed well at distinguishing between ADRs and non-ADRs when validated on the OMOP NSA reference set (AUC ranging between 0.792 and 0.940). The classifier's performance was better than existing unsupervised methods' performances calculated in previous studies [39] highlighting the advantages of implementing supervised learning for signalling ADRs.

Suggested future areas of work are expanding the feature engineering to include the remaining Bradford Hill causality considerations (analogy, consistency, plausibility and coherence) and using the Map-Reduce paradigm to enable the extraction of the features in quasi-linear scalability, making this framework suitable for terabytes of big healthcare data.

Appendix A. Feature Summaries

Table A.7: Summary of the features used within the SADR framework.

| Feature | Category | Description |
|---|---|---|
| $Attr_1$ | Strength | The fraction of patients experiencing the health outcome within 30 days of the drug minus the fraction of patients experiencing the health outcome within 30 days of any other drug |
| $Attr_2$ | Strength | The fraction of patients experiencing the health outcome within 30 days of any first time prescription in 13 months of the drug minus the fraction of patients experiencing the health outcome within 30 days of a first time prescription in 13 months of any other drug |
| $Attr_3$ | Strength | The fraction of patients experiencing the health outcome within 30 days of any first time prescription of a drug family minus the fraction of patients experiencing the health outcome within 30 days of a first time prescription of any other drug family. |
| $Attr_4$ | Specificity | The average age of patients experiencing the health outcome within 30 days of the drug divided by the average age of patients prescribed the drug |
| $Attr_5$ | Specificity | The average age of patients experiencing the health outcome within 30 days of any first time prescription in 13 months of the drug divided by the average age of patients first time prescriptions in 13 months of the drug. |
| $Attr_6$ | Specificity | The average age of patients experiencing the health outcome within 30 days of any first time prescription of the drug family divided by the average age of patients first time prescription of the drug family |
| $Attr_7$ | Specificity | The fraction of patients that are male out of those experiencing the health outcome within 30 days of the drug divided by the male fraction of patients prescribed the drug |
| $Attr_8$ | Specificity | The fraction of patients that are male out of those experiencing the health outcome within 30 days of any first time prescription in 13 months of the drug divided by the fraction of patients prescriptions corresponding to the first time prescription in 13 months of the drug where the patient is male |
| $Attr_9$ | Specificity | The fraction of patients that are male out of those experiencing the health outcome within 30 days of any first time prescription of the drug family divided by the fraction of patients prescriptions corresponding to the first time prescription of the drug family where the patient is male |
| $Attr_{10}$ | Specificity | The hierarchal level of the Read code (this corresponds to how specific the health outcome is) |
| $Attr_{11}$ | Temporality | The number of times the heath outcome occurs within 1 and 30 days after a prescription of the drug divided by the number of times the heath outcome occurs within 1 and 30 days before a prescription of the drug |
| $Attr_{12}$ | Temporality | The number of times the heath outcome occurs within 1 and 30 days after a first time in 13 months prescription of the drug divided by the number of times the heath outcome occurs within 1 and 30 days before a first time in 13 months prescription of the drug |
| $Attr_{13}$ | Temporality | The number of times the heath outcome occurs within 1 and 30 days after a first time prescription of the drug family divided by the number of times the heath outcome occurs within 1 and 30 days before a first time prescription of the drug family |
| $Attr_{14}$ | Biological gradient | The average drug dosage given to patients experiencing the health outcome within 30 days of the drug divided by the average drug dosage given to patients prescribed the drug |
| $Attr_{15}$ | Biological gradient | The average drug dosage given to patients experiencing the health outcome within 30 days of any first time prescription in 13 months of the drug divided by the average drug dosage given to patients when considering their first time prescription in 13 months of the drug. |
| $Attr_{16}$ | Biological gradient | The average drug dosage given to patients experiencing the health outcome within 30 days of any first time prescription of the drug family divided by the average drug dosage given to patients when considering their first time prescription of the drug family |
| $Attr_{17}$ | Experimentation | The number of patients who were prescribe the drug and experienced the health outcome |
| | | within 1 and 30 days after on two or more distinct time periods (with 13 months or more between consecutive prescriptions of the drug) but never experienced the health outcome within 1 and 30 days prior to any prescription of the drug divided by the number of patients with multiple distinct prescription periods (with two or more drug prescriptions with a gap of 13 months of more between consecutive prescriptions). |
| $Attr_{18}$ | Hierarchal | The average number of other drugs prescribed during the 30 days before and after the drug prescription for the prescriptions where the patient experienced the health outcome within 1 and 30 days after the prescription divided by the average number of other drugs prescribed during the 30 days before and after the drug prescription for all prescriptions of the drug. |
| $Attr_{19}$ | Hierarchal | The average number of other drugs prescribed during the 30 days before and after any first time in 13 month prescription of the drug where the patient experienced the health outcome within 1 and 30 days after the prescription divided by the average number of other drugs prescribed during the 30 days before and after the drug prescription for any first time in 13 months prescription of the drug. |
| $Attr_{20}$ | Hierarchal | The average number of other drugs prescribed during the 30 days before and after any first time prescription of the drug family where the patient experienced the health outcome within 1 and 30 days after the prescription divided by the average number of other drugs prescribed during the 30 days before and after the drug prescription for any first time prescription of the drug family. |
| $Attr_{21}$ | Hierarchal | The number of prescriptions where the patient has the health outcome recorded within 1 and 30 days after the drug but the patient has never had a similar but more general version of the health outcome recorded (level 4 version read code) divided by the number of prescriptions where the patient has the health outcome recorded within 1 and 30 days after the drug but the patient has never had a similar but even more general version of the health outcome recorded (level 3 version read code). |
| $Attr_{22}$ | Hierarchal | The number of prescriptions where the patient has the more general level 4 version of the health outcome recorded within 1 and 30 days after the drug prescription divided by the number of prescriptions where the patient has the more general level 4 version of the health outcome recorded within 1 and 30 days before the drug prescription. |
| $Attr_{23}$ | Hierarchal | The number of prescriptions where the patient has the more general level 3 version of the health outcome recorded within 1 and 30 days after the drug prescription divided by the number of prescriptions where the patient has the more general level 3 version of the health outcome recorded within 1 and 30 days before the drug prescription. |
| $Attr_{24}$ | Hierarchal | The number of prescriptions where the patient has the more general level 4 version of the health outcome recorded within 1 and 30 days after the first time in 13 month prescription of the drug divided by the number of prescriptions where the patient has the more general level 4 version of the health outcome recorded within 1 and 30 days before the first time in 13 months prescription of the drug. |
| $Attr_{25}$ | Hierarchal | The number of prescriptions where the patient has the more general level 3 version of the health outcome recorded within 1 and 30 days after the first time in 13 month prescription of the drug divided by the number of prescriptions where the patient has the more general level 3 version of the health outcome recorded within 1 and 30 days before the first time in 13 months prescription of the drug. |
| $Attr_{26}$ | Hierarchal | The number of prescriptions where the patient has the more general level 4 version of the health outcome recorded within 1 and 30 days after the first time prescription of the drug family divided by the number of prescriptions where the patient has the more general level 4 version of the health outcome recorded within 1 and 30 days before the first time prescription of the drug family. |
| $Attr_{27}$ | Hierarchal | The number of prescriptions where the patient has the more general level 3 version of the health outcome recorded within 1 and 30 days after the first time prescription of the drug family divided by the number of prescriptions where the patient has the more general level 3 version of the health outcome recorded within 1 and 30 days before the first time prescription of the drug family. |